\documentclass[10pt, a4paper]{article}

\usepackage[final]{lrec2026} 
\usepackage{mdframed}
\usepackage{booktabs}
\usepackage{multirow}
\usepackage{graphicx}
\usepackage{tabularx}
\usepackage{array}
\usepackage{appendix}
\usepackage{caption}
\usepackage{xcolor}
\usepackage{cuted} 
\usepackage{caption}
\usepackage{placeins}
\usepackage[utf8]{inputenc}
\newcolumntype{Y}{>{\raggedleft\arraybackslash}X}
\newcolumntype{C}{>{\centering\arraybackslash}X}
\newcolumntype{Z}{>{\raggedright\arraybackslash}X}
\usepackage{ragged2e}

\usepackage{listings}
\usepackage{enumitem}

\title{Estonian WinoGrande Dataset: Comparative Analysis of LLM Performance on Human and Machine Translation}

\name{Marii Ojastu$^{1,2,*}$, Hele-Andra Kuulmets$^{1}$, Aleksei Dorkin$^{1}$, \\ 
\bfseries \large Marika Borovikova$^{2}$, Dage Särg$^{3}$, Kairit Sirts$^{1,*}$} 

\address{$^{1}$TartuNLP, Institute of Computer Science\\
         $^{2}$Department of Translation Studies, Institute of Foreign Language and Cultures\\
         $^{3}$Institute of Genomics\\
         University of Tartu, Estonia\\
         \{firstname.lastname\}@ut.ee\\}

\abstract{In this paper, we present a localized and culturally adapted Estonian translation of the test set from the widely used commonsense reasoning benchmark, WinoGrande. We detail the translation and adaptation process carried out by translation specialists and evaluate the performance of both proprietary and open source models on the human translated benchmark. Additionally, we explore the feasibility of achieving high-quality machine translation by incorporating insights from the manual translation process into the design of a detailed prompt. This prompt is specifically tailored to address both the linguistic characteristics of Estonian and the unique translation challenges posed by the WinoGrande dataset. Our findings show that model performance on the human translated Estonian dataset is slightly lower than on the original English test set, while performance on machine-translated data is notably worse. Additionally, our experiments indicate that prompt engineering offers limited improvement in translation quality or model accuracy, and highlight the importance of involving language specialists in dataset translation and adaptation to ensure reliable and interpretable evaluations of language competency and reasoning in large language models.
\\ \newline \Keywords{WinoGrande, benchmark translation, Estonian}
}

\begin{document}

\maketitleabstract

\section{Introduction}

The WinoGrande dataset~\citep{sakaguchi2019winograndeadversarialwinogradschema} is a widely used benchmark for evaluating the commonsense reasoning abilities of large language models (LLMs)~\citep{team2024gemma,jiang2024mixtral,touvron2023llama,grattafiori2024llama}. However, this dataset is exclusively in English, limiting evaluation of commonsense reasoning only to that language. Meanwhile, large language models have recently become increasingly multilingual, creating the need for comparable benchmarks that assess these capabilities beyond English.

A common approach to creating benchmarks for other languages is to translate an existing benchmark from English to the target language. To reduce translation costs, machine translation (MT) systems are often used for this purpose ~\citep[]{lai-etal-2023-okapi,  foroutan-etal-2023-breaking, thellmann2024multilingualllmevaluationeuropean, singh-etal-2024-aya, dang2024aya, raihan-etal-2025-mhumaneval}. However, evaluation results on machine-translated data tend to differ from those on human-annotated data \citep{kreutzer2025d}. This difference is not easily predictable and may arise from multiple factors such as models exploiting translation artifacts \citep{artetxe-etal-2020-translation} or translation inaccuracies in test data that obscure the benchmark task \citep{plaza2024spanishllmbenchmarksmmlu}.
Moreover, machine translation lacks mechanisms for localization or cultural adaptation, resulting in datasets that are semantically distant from natural language use by native speakers. Finally, the WinoGrande benchmark is particularly difficult to translate across languages due to the inherent nature of the task and the strict constraints each example must satisfy.

\begin{table}[!t]
\small
\centering
\begin{tabularx}{\columnwidth}{lX}
\toprule
\textbf{Schema} & The door opened louder than the window because the \_ had less grease on its hinges. \\
\midrule
\textbf{Option 1} & window \\
\textbf{Option 2} & door \\
\bottomrule
\end{tabularx}
\caption{An example from the WinoGrande dataset.}
\label{tab:schema_example}
\end{table}

Considering the aforementioned aspects, we manually translated, localized and culturally adapted the WinoGrande test set, which consists of 1,767 instances, into Estonian, a mid-resource Finno-Ugric language, to support development and evaluation of language models for this language. We discuss in detail the linguistic and translation challenges encountered in the process, including the correction of ambiguous and incorrectly labelled examples that were identified during this work.

We compared this newly created manually translated dataset to two machine translated versions produced by OpenAI models. Our experiments with open and proprietary models show higher accuracy on the manually translated, localized, adapted and corrected WinoGrande than on the machine-translated versions, with gains in both the localized-adapted and corrected subsets.

To further examine the differences, we manually identified machine translated examples where the meaning has shifted from the original meaning. We observe that, on average, models score lower on these examples than on human-translated equivalents, indicating potential label–sentence inconsistencies introduced by machine translation.

Among the 1,767 examples in the translated WinoGrande test set, we manually corrected 89 (5\% of the entire test set), while localization and adaptation was applied to 53 examples (3\%). 
Among the remaining semantically comparable samples, up to 15.2\% lost their original semantics during machine translation. 
Taken together, these results demonstrate the ongoing infeasibility of using current state-of-the-art LLMs to reliably translate WinoGrande sentences into Estonian.\footnote{The human translated Estonian WinoGrande dataset is available at \url{https://huggingface.co/datasets/tartuNLP/winogrande_et}}

\section{WinoGrande Dataset}

The Winograd Schema Challenge (WSC) \citep{winograd} is a pronoun reference disambiguation task proposed to evaluate commonsense knowledge in AI systems and has inspired a range of subsequent benchmarks \citep[]{sakaguchi2019winograndeadversarialwinogradschema,rudinger2018genderbiascoreferenceresolution,zhang-etal-2020-winowhy} for testing AI capabilities. 
The original WSC is a set of few hundred expert-crafted examples intended only to assess AI capabilities. However, it was later demonstrated to be solvable with fine-tuning deep neural networks on auxiliary datasets without particular advancements in commonsense reasoning being made \citep[][]{kocijan-etal-2019-surprisingly, sakaguchi2019winograndeadversarialwinogradschema}.

To address this issue, \citet{sakaguchi2019winograndeadversarialwinogradschema} introduced WinoGrande, a large-scale ambiguity resolution dataset, in total consisting of 44k crowdsourced problems inspired by the original Winograd Schema Challenge.
The dataset consists of sentences where in the first part of a sentence two nouns are mentioned and the second part contains a blank which corresponds to the mention of one of the nouns (see Table \ref{tab:schema_example}). Some sentences are paired (hereafter twin sentences), have a lexical overlap on 70\% and share the same set of answer options (see Table \ref{tab:example}). The 
objective is to decide which of the two nouns correctly fills the blank. 

According to \citet{sakaguchi2019winograndeadversarialwinogradschema}, the sentences are designed to meet two additional criteria. First, the answer options must be unambiguous, meaning that humans would easily be able to select the correct option without considerable effort. The authors report that human performance on the test set is 94\%. Secondly, the correct answer option should not be derivable solely from the local context or simple word associations. To enforce the second constraint, the authors applied a bias-reduction algorithm that extends the idea of filtering out human-detectable lexical biases to the embedding space. The algorithm identifies and removes examples that a model could solve by exploiting statistical regularities in word embeddings rather than true reasoning. Despite extensive manual and automatic filtering, the dataset has been reported to have several issues---such as instances solvable through simple word correlations, poorly written examples, sentences revealing the correct answer, or examples that are genuinely difficult to understand \citep{kocijan2023defeatwinogradschemachallenge}---and is not completely free of artifacts \citep{elazar-etal-2021-back}.

\begin{table}[t]
\centering
\small
\setlength{\tabcolsep}{4.5pt}
\begin{tabularx}{\columnwidth}{ll}
\toprule
\textbf{WSC translations} & \textbf{Type} \\
\midrule

French \cite{amsili-seminck-2017-google} & HT \\
Portuguese \cite{eniac} & HT \\
Japanese \cite{shibata2015nihongo}& HT  \\
Chinese \cite{bernard-han-2020-mandarinograd} & HT \\
Hungarian \cite{Vadsz2022WinogradSA} & MT+PE \\
Thai \cite{artkaew-2025-thai} & HT  \\
Hebrew \cite{shwartz2024winograd} & N/A \\

\midrule
\textbf{WG translations} & \textbf{Type} \\
\midrule

Icelandic \cite{snaebjarnarson-etal-2022-warm} & HT \\
8 African languages \cite{Alhanai_2025} & HT \\
Arabic + dialects \cite{mousi-etal-2025-aradice} & MT+PE \\
Italian \cite{moroni-etal-2024-towards} & MT \\
Kyrgys \cite{11206960} & MT+PE \\
Nepali \cite{nyachhyon2025consolidatingdevelopingbenchmarkingdatasets} & MT+PE \\
Egyptian \cite{shang2025nilechategyptianlanguagemodels} & MT \\
Romanian \cite{masala-etal-2024-vorbesti} & MT \\
Lithuanian \cite{Nakvosas_2025} & MT \\
Korean \cite{kim2025openkollmleaderboard2bridging} & HT \\
Basque* \cite{corral2024pipelineanalysisdevelopinginstruct} & HT \\
Persian \cite{farsi2025melacmassiveevaluationlarge} & MT+PE \\

\bottomrule
\end{tabularx}
\caption{Overview of Winograd Schema Challenge (WSC) and WinoGrande (WG) dataset translations across languages. HT indicates human translation, MT indicates machine translation, and PE indicates post-editing. Asterisk (*) marks a translation of a subset of samples.}
\label{tab:other_translations}
\end{table}

\begin{table*}[t]
\centering
\small
\begin{tabularx}{\linewidth}{p{0.1cm}p{4cm}p{4cm} X}
    \toprule
    \bf Ex. & \bf English Schema & \bf Estonian Schema & \bf Comment\\
    \midrule
     (A) & I much prefer the necklace that I have over the bracelet of my friend because the \_ is cheap. \newline \textbf{Option 1}: necklace \newline \textbf{Option 2}: bracelet & Ma eelistan oma sõbra käevõru asemel pigem oma kaelakeed, sest see \_ on maitsetu. \newline \textbf{Option 1}: kaelakee \newline \textbf{Option 2}:  käevõru & This example shows the need for deliberate choices to avoid ambiguity. In English, \emph{cheap} implies both low price and low quality, while its Estonian equivalent \emph{odav} refers mainly to low price. In translation, \emph{cheap} is rendered as \emph{maitsetu} (``tacky'') rather than literally, preserving the schema’s intended clarity and disambiguation.\\
     \midrule
     (B) & The girl ate less beans than meat on the date because the \_ made her full. \newline \textbf{Option 1}: beans \newline \textbf{Option 2}: meat & \textbf{Human translated:} \newline Tüdruk sõi kohtingul ube vähem kui liha, sest tal sai \_ kõht täis. \newline \textbf{Option 1}: ubadest \newline \textbf{Option 2}: lihast
     \par\smallskip
     \textbf{Machine translated:} \newline Tüdruk sõi kohtingul vähem ube kui liha, sest \_ tegi ta kõhu täis. \newline \textbf{Option 1}: oad \newline \textbf{Option 2}: liha & The machine translation introduces a number mismatch: in English, \emph{beans} (plural) and \emph{meat} (singular) pose no issue, but in Estonian, the verb \emph{tegi} (``made'') agrees only with the singular. This allows the model to rely on grammar rather than reasoning, since the correct answer is also the only grammatically fitting one. The human translation fixes this by rephrasing the schema as \emph{became full from the \_} instead of the \par literal \emph{\_ made her full}, removing the grammatical cue.\\
    \bottomrule
\end{tabularx}
\caption{Examples of translated WinoGrande schemas.}
\label{tab:translation_examples}
\end{table*}

Although the WinoGrande dataset has turned out to be less challenging for large language models than anticipated \citep{Lourie_Le}, it measures the progress only for English, leaving it unclear how well the success of solving ambiguity resolution tasks transfers to other languages. To answer this question, both the Winograd Schema Challenge and WinoGrande have been translated into several other languages (Table \ref{tab:other_translations}). Additional translations exist \citep{shavrina-etal-2020-russiansuperglue,aparovich-etal-2025-belarusianglue, zagar-robnik-sikonja-2022-slovene}, for a recast version of the Winograd Schema Challenge dataset that is part of the SuperGLUE benchmark \cite{wang2020supergluestickierbenchmarkgeneralpurpose}. However, as \citet{amsili-seminck-2017-google} noted, translating these tasks is not a straightforward process and requires solving several linguistic challenges such as the disagreement of translated nouns in number or gender or general ambiguity of the translated schema.

Another issue with translating these datasets to other languages is the culture- and region-specific knowledge that some of these examples assume and which might be in the role of commonsense knowledge that is needed to resolve ambiguity. However, for the speakers of non-English languages, this type of knowledge can not be expected to be part of commonsense knowledge meaning that such examples must be specifically adapted for the target language. For instance, \citet{snaebjarnarson-etal-2022-warm} reports having to do cultural adaptation of some of the examples. \citet{Alhanai_2025} assessed cultural appropriateness of WinoGrande translations into 11 African languages and found that 20.6\% examples can be considered culturally inappropriate.

\section{Human Translated Dataset}
This section presents the steps involved in creating a human-translated Estonian version of the WinoGrande dataset. It first describes the translation process, followed by the description of localization, cultural adaptation, and error correction. Finally, to ensure translation quality, we report inter-annotator agreement on this newly created dataset.

\subsection{Translation Process}
The translation of the WinoGrande test set into Estonian was carried out by a master's student in translation studies as part of their thesis work. The translations were revised in collaboration with a professional translator, who is a master's level expert in translation studies. Both translators are co-authors of this paper.

The translation aimed to preserve as many features of the original dataset as possible. In case of twin sentences, a 70\% lexical overlap was maintained between the sentences, following \citet{sakaguchi2019winograndeadversarialwinogradschema}. Furthermore, since both sentences in a pair were required to share the same set of answer options, which were identical not only in lexical form but also in grammatical case, deliberate manual adjustments were necessary in the translation. Notably, achieving this is more difficult in Estonian, as its agglutinative structure complicates the preservation of morphological uniformity. 

When a direct translation resulted in ambiguous schemas---often because the Estonian equivalent of a word had a broader or narrower meaning than its English counterpart---problematic words were substituted with alternatives that preserved the intended objective of the schema, similarly to \citet{amsili-seminck-2017-google}. An example of this type of adjustment is presented in Table \ref{tab:translation_examples} (A).

Similarly to  \citet{amsili-seminck-2017-google}, adjustments were made in cases where a direct translation of the nouns would have resulted in disagreement in number and would have made the instance trivial to solve based on verb morphology. An example of such adjustment is presented in Table~\ref{tab:translation_examples}~(B).

Finally, in order to allow models' evaluation in the few-shot setting, we also translated six schemas from the development set. We selected instances that represent the variability of the tasks, for instance, sentences from the social and physical domain and single and twin sentences.

\subsection{Localization, Adaptation and Error Correction}
The dataset was localized by replacing geographical locations, brands, foods, activities, and animal species with culturally and regionally appropriate Estonian equivalents. A total of 53 samples were localized. As a result of the localization process, two types of samples can be identified: those in which the culturally adapted information is necessary for reasoning ( Table \ref{tab:cultural_adaptation_examples}, A) , and those in which it is not ( Table \ref{tab:cultural_adaptation_examples}, B). Names were adapted in all samples, 
with English names replaced by typical Estonian names.

\begin{table}[t]
\centering
\small
\begin{tabularx}{\columnwidth}{p{3.4cm} X}
    \toprule
    \bf English Schema & \bf Culturally Adapted Estonian Schema \\
    \midrule
    
    (A) Samantha lived in the \underline{city} while Patricia lived in the \underline{desert}, so \_ thought it common to see a \underline{cactus} in their yard. 
    \newline \textbf{Option 1}: Samantha 
    \newline \textbf{Option 2}: Patricia 
    & 
    Sandra elas \underline{mandril}, kuid Piret elas \underline{saarel}. Seega oli \_ jaoks oma aias \underline{kadaka} nägemine tavapärane. 
    \newline \textbf{Option 1}: Sandra 
    \newline \textbf{Option 2}: Piret \\
    
    \midrule
    
    (B) Laura beat Erin in a game of \underline{Mortal Kombat}, then \_ said congratulations on the win. 
    \newline \textbf{Option 1}: Laura 
    \newline \textbf{Option 2}: Erin 
    & 
    Laura tegi Erele lauamängus \underline{„Mees,} \underline{kes teadis ussisõnu“} pähe, mispeale õnnitles \_ teda võidu puhul. 
    \newline \textbf{Option 1}: Laura 
    \newline \textbf{Option 2}: Ere \\
    
    \bottomrule
\end{tabularx}
\caption{Examples of culturally adapted WinoGrande schemas in English and Estonian. In Sample A, correct resolution requires cultural knowledge: the original contrasts city and desert (e.g., cacti in deserts), while the Estonian version replaces these with mainland (\emph {mandril}) and island (\emph{saarel}), and cactus with juniper (\emph{kadaka}), a plant associated with Estonian islands. In Sample B, the game name is culturally adapted, but this knowledge is not necessary for reasoning.}
\label{tab:cultural_adaptation_examples}
\end{table}

\begin{table*}[ht]
\centering
\small
\begin{tabularx}{\textwidth}{p{3cm}XX}
\toprule
\textbf{Language} & \textbf{Twin Sentence 1} & \textbf{Twin Sentence 2} \\
\midrule
English & I spilled water and jello on the buttons of the remote. It being sticky is probably \underline{not} because of the \_.  
\newline \textbf{Option 1}: jello \newline \textbf{Option 2}: water 
& I spilled water and jello on the buttons of the remote. It's being sticky probably because of the \_.  
\newline \textbf{Option 2}: jello \newline \textbf{Option 2}: water \\
\addlinespace
Estonian \newline Machine translation \newline Simple prompt & Ma ajasin vett ja \underline{želeed pultnuppudele.} \underline{Selle kleepuv olemine ei ole} tõenäoliselt \underline{tingitud \_}.  
\newline \textbf{Option 1}: \underline{želee} \newline \textbf{Option 2}: vesi 
& Ma ajasin vett ja \underline{tarretist puldi nuppudele.} \underline{See on} tõenäoliselt \underline{kleepuv \_ tõttu}.  
\newline \textbf{Option 1}: \underline{tarretis} \newline \textbf{Option 2}: vesi \\
\addlinespace
Estonian \newline Machine translation \newline Detailed prompt & Ma valasin vett ja želeed puldinuppudele. \underline{Selle kleepuvus ei ole} tõenäoliselt \_ tõttu.  
\newline \textbf{Option 1}: želee \newline  \textbf{Option 2}: vee
& Ma valasin vett ja želeed puldinuppudele. \underline{See on} tõenäoliselt \underline{kleepuv} \_ tõttu.  
\newline \textbf{Option 1}: želee \newline \textbf{Option 2}: vee \\
\addlinespace
Estonian \newline Human translation & Ma ajasin puldi nuppudele vett ja tarretist. Selle kleepumine \underline{ei tulene} ilmselt \_.  
\newline \textbf{Option 1}: tarretisest \newline \textbf{Option 2}: veest 
& Ma ajasin puldi nuppudele vett ja tarretist. Selle kleepumine \underline{tuleneb} ilmselt \_.  
\newline \textbf{Option 1}: tarretisest \newline \textbf{Option 2}: veest \\
\bottomrule
\end{tabularx}
\caption{The machine-translated (Simple prompt) twin sentences show substantial lexical divergence (marked with underline). Additionally, the answer options differ across the two sentences. Finally, the options are rendered in the nominative case, which does not fit the grammatical context of the sentence. The machine translation using the detailed prompt results in less lexical divergence between twin sentences and ensures that their corresponding options remain consistent with the intended format, with each pair of twin sentences being assigned identical options. The human translation is also given for the reference.}
\label{tab:example}
\end{table*}

During the manual translation process, 89 samples in the original dataset were found to be either inherently ambiguous or had the wrong answer marked as correct. Such issues were addressed in the human translated dataset by adjusting the schema or editing its answer options to eliminate ambiguity and ensure correctness. 
The labels marking those problematic items in the original WinoGrande dataset will be released together with the Estonian WinoGrande dataset.

\subsection{Inter-Annotator Agreement}
The human translation and adaptation was carried out so that the original answer options were always retained. 
For example, if the correct answer in the original English schema was option 1, the translated and adapted version preserved the same correct option.
To assess human-level agreement on the resulting dataset, two additional annotators (both co-authors) independently labeled all items in the translated dataset. Annotators were instructed to select the correct answer option if it was clearly inferable from the context, or to mark it "undecidable" if the item appeared ambiguous and the correct answer could not be determined.

The Cohen's kappa between the two annotators is 0.816, and the Fleiss' kappa between all three annotations (including the original) was 0.855, both values falling into the very high agreement range according to standard interpretation conventions \citep{landis1977measurement}. The accuracy was 95\% for Annotator 1 and 92\% for Annotator 2. Annotator 1 found 22 items undecidable, while Annotator 2 labeled 106 tasks as such. In the majority of the cases, the annotators disagreed on which items were undecidable, with the overlap being only in eight cases.\footnote{These samples are marked in our dataset.}

\section{Machine Translated Dataset}
We compare the human-translated dataset with two machine-translated versions produced using GPT-4o and GPT-4.1. The first version was generated with a short, generic translation prompt, while the second was produced using longer, detailed prompts designed to address the issues observed in the first machine translated version. 

\subsection{Simple Prompt}
After producing the translations with the simple zero-shot translation prompt (\ref{subsec:simple_p}), we manually analyzed them for any systematic issues that could hinder the objective of the tasks or result in deviations from the required format, thereby potentially affecting the benchmark results.

The manual analysis of the machine translated data revealed the existence of translated schemas that can be solved based on grammatical cues rather than commonsense reasoning (see an example in Table \ref{tab:translation_examples} (B)). This can lead to predictions that are either incorrect or correct for the wrong reasons \citep{elazar-etal-2021-back,mccoy2019right}, making the schema unreliable for its intended purpose. Additionally, machine translation sometimes introduced ambiguity into the schema (see an example in Table~\ref{tab:translation_examples}~(A)).

In WinoGrande dataset, the answer options always appear in the sentence. What distinguishes the English dataset from the Estonian dataset is the way the answers appear in the schema. In the English dataset, the answer and its antecedent typically appear in the same form, except for changes in the possessive case for antecedents. In contrast, in the Estonian dataset both the answer and its antecedent can appear in any of the 14 grammatical cases, with the case of the antecedent usually being different from the case of the answer.
This nuance becomes apparent in the machine-translated dataset, where the answer options are translated independently of the sentence context and are usually rendered in the nominative case. Consequently, they often fail to fit grammatically within the sentence, which may require a different case for proper syntactic agreement (see Table~\ref{tab:example}).

For twin sentences, we observed that  machine translation often fails to maintain structural and lexical consistency. The machine translated sentence pairs frequently diverge in wording, and the answer choices may be translated inconsistently (see Table~\ref{tab:example}). While this misalignment may not affect overall performance metrics, it undermines the integrity of the dataset’s original format.

\subsection{Detailed Prompt}

Next, we developed two detailed prompts---one for single sentences (\ref{subsec:detailed_p}) and the other for the twin sentences (\ref{subsec:detailed_twin_p})---and used them to generate a revised translation of the dataset using OpenAI GPT-4.1. The prompts were supplemented with five single and ten twin few-shot examples, respectively. The additional samples were manually selected and translated to reflect different linguistic challenges in the task translation.

These prompts specifically targeted systematic shortcomings identified in the initial machine translation with the simple prompt, such as incorrect inflection of answer options, insufficient lexical overlap between twin sentences (less than 70\%), numerical mismatches, and overly literal translations that could compromise the interpretability of the schema. The prompt was supplemented with a list of Estonian names, each labeled with gender (male or female) and inflected across all 14 grammatical cases, to enable automatic localization of names in the translation. The instruction for substituting the names with Estonian ones from the list was included in the prompt.

\section{Benchmarking Models}

Our comparison of the newly created datasets is based on their applicability as reliable benchmarks for evaluating commonsense reasoning in Estonian. For that purpose, we first used them to obtain and compare overall results across a range of open and closed LLMs (this section). We then conducted a subset-level analysis (Section \ref{sec:subset_analysis}) of these results to better understand the differences in results across the datasets. 

The set of models evaluated is following: 

\begin{itemize}
    \item Five proprietary models: Gemini 2.5 Pro, Gemini 2.5 Flash~\cite{gemini2025gemini25}, Claude Sonnet 4.5~\cite{anthropic2025claude45sonnet}, GPT-4.1 \cite{openai2024gpt4technicalreport}, GPT-5 \cite{GPT-5};
    \item Four open models in the moderate to large range: Llama~3.3-70B and Llama~3.1-405B~\cite{llama3modelcard}, Gemma~3-27B 
\cite{gemma2025gemma3},
and Qwen~2.5-72B \cite{yang2025qwen3technicalreport};
    \item  Three open models in the smaller range: Llama~3.1-8B ~\cite{llama3modelcard}, Apertus~8B \cite{hernandezcano2025apertusdemocratizingopencompliant}, EuroLLM~9B \cite{martins2024eurollmmultilinguallanguagemodels}.
\end{itemize}

All models were evaluated on all four versions of the WinoGrande dataset: the original English dataset, the human translated Estonian WinoGrande, and the two machine-translated Estonian versions of the dataset. The Gemini 2.5 Pro model was evaluated with minimum thinking (128 token budget), Gemini 2.5 Flash and GPT-5 were evaluated in no thinking mode, and Claude Sonnet 4.5 was evaluated in default thinking mode. 
All models were prompted in few-shot setting using three examples from the translated development set items.
The accuracy results are presented in Table  \ref{tab:model_results}. 

\begin{table}[t]
    \centering
    \small
    \setlength{\tabcolsep}{0.5em}
    \begin{tabular}{lrrrr}
        \toprule
        & \multirow{2}{*}{\bf EN} & \multirow{2}{*}{\bf HT} & \multicolumn{2}{c}{\bf MT} \\
        \cmidrule(lr{0.2em}){4-5}
        & & & \multicolumn{1}{c}{\bf Simple} & \multicolumn{1}{c}{\bf Detailed} \\
        \midrule
        \multicolumn{5}{l}{\textbf{Proprietary models}} \\
        \midrule
         Gemini 2.5 Pro & 89.6 & 91.1 & 83.9 & 82.7 \\
         Gemini 2.5 Flash & 87.8 & 88.6 & 82.2 & 81.8 \\
         Claude Sonnet 4.5 & 94.7 & 93.7 & 88.3 & 87.4 \\
         GPT-4.1 & 85.4 & 82.6 & 76.1 & 77.2 \\
         GPT-5 & 83.5 & 82.9 & 76.6 & 76.4 \\
         \cmidrule(lr){1-5}
         Average & 88.2 & 87.8 & 81.4 & 81.1\\
         \midrule
         \midrule
         \multicolumn{5}{l}{\textbf{Open models: moderate to large range}} \\
         \midrule
         Gemma 3-27B & 73.6 & 74.6 & 68.8 & 69.6\\
         Qwen 2.5-72B & 83.4 & 71.4 & 65.7 & 66.2 \\
         Llama 3.3-70B & 79.9 & 74.8 & 69.4 & 69.7 \\
         Llama 3.1-405B & 84.3 & 78.4 & 72.0 & 72.7 \\
         \cmidrule(lr){1-5}
         Average & 80.3 & 74.8 & 68.9 & 69.6\\
         \midrule
         \midrule
         \multicolumn{5}{l}{\textbf{Open models: smaller range}} \\
         \midrule
         Llama 3.1-8B & 55.5 & 54.2 & 53.8 & 53.1 \\
         Apertus-8B & 51.9 & 51.5 & 51.2 & 51.2 \\
         EuroLLM-9B & 59.6 & 61.3 & 59.4 & 57.7\\
         \cmidrule(lr){1-5}
         Average & 55.7 & 55.7 & 54.8 & 54.0\\
         \bottomrule
    \end{tabular}
    \caption{Accuracy results reported across models for the WinoGrande English test set (EN), Estonian human-translated version (HT), the Estonian machine-translated version using a simple prompt (MT Simple), and the detailed prompt (MT Detailed)}
    \label{tab:model_results}
\end{table}

Compared to the original English WinoGrande test set, the performance of proprietary models on the Estonian human-translated dataset was generally similar, with the average difference being only 0.4\%.

Performance differences were more pronounced among open models in the moderate to large model range, where on average, the models performed 5.5\% better on English compared to human translated Estonian dataset. While most models performed worse on the Estonian dataset, interestingly Gemma 3-27B performed slightly better.

Among smaller models, the average performance was the same on both English and Estonian datasets, with Llama 3.1-8B and Apertus-8B performing only slightly above chance level on both datasets. EuroLLM-9B was the strongest of the small models according to accuracy, and performed 1.7\% better on the Estonian dataset than on the English one.

Comparing the machine translated Estonian WinoGrande versions to the human translated version reveals that all models obtain higher performance on the human translated dataset. A detailed investigation of this observation is presented in Section~\ref{sec:subset_analysis}.

Contrasting the results on the two machine translation versions reveals only minor differences, suggesting that the translations obtained with the detailed prompt did not yield an substantial effect on the models' performance.

\section{Subset-Level Analysis} \label{sec:subset_analysis}
We analyzed the impact of cultural adaptation, and error correction in human translation on the overall results by calculating the models' accuracies for subsets of samples that were culturally adapted (n = 53), corrected (n = 89), and the rest that were deemed semantically comparable (n = 1,617). The 8 samples that both annotators found ambiguous were excluded from the subset analysis.
For these analyses, we compared the results of open models of moderate to large size, as smaller models often performed near random even on the English dataset. We compared the results across three versions of the dataset---the original English, the human translated Estonian, and the machine translated Estonian created with the detailed prompt.\footnote{The machine translated dataset created with the simple prompt shows very similar results.}

\begin{table}[t]
    \centering
    \begin{tabular}{lrrr}
        \toprule
        & \bf EN & \bf HT & \bf MT \\
        \midrule
        \multicolumn{4}{l}{\textbf{Culturally adapted } $n = 53$} \\
        \cmidrule(lr){1-4}
         Gemma 3-27B & 84.9 & 81.1 & 77.8 \\
         Qwen 2.5-72B & 86.8 & 77.4 & 73.6 \\
         Llama 3.3-70B & 88.7 & 81.1 & 81.5 \\
         Llama 3.1-405B & 92.5 & 83.0 & 81.1 \\
         \cmidrule(lr){1-4}
         Average & 88.2 & 80.7 & 78.5 \\
         \midrule
         \midrule
         \multicolumn{4}{l}{\textbf{Corrected } $n = 89$} \\
         \cmidrule(lr){1-4}
         Gemma 3-27B & 49.4 & 71.9 & 41.6 \\
         Qwen 2.5-72B & 58.4 & 68.5 & 42.7 \\
         Llama 3.3-70B & 61.8 & 73.0 & 47.2 \\
         Llama 3.1-405B & 56.2 & 78.7 & 48.3 \\
         \cmidrule(lr){1-4}
         Average & 56.5 & 73.0 & 44.9 \\
         \midrule
         \midrule
         \multicolumn{4}{l}{\textbf{Semantically comparable } $n = 1,617$} \\
         \cmidrule(lr){1-4}
         Gemma 3-27B & 74.5 & 74.6 & 70.9 \\
         Qwen 2.5-72B & 84.6 & 71.2 & 67.3 \\
         Llama 3.3-70B & 80.5 & 74.6 & 70.4 \\
         Llama 3.1-405B & 85.6 & 78.2 & 74.0 \\
         \cmidrule(lr){1-4}
         Average & 81.3 & 74.7 & 70.7 \\
         \bottomrule
    \end{tabular}
    \caption{Accuracy on different subsets for the English (EN), Estonian human-translated (HT), and the Estonian machine-translated version using the detailed prompt (MT).
    }
    \label{tab:subsets_results}
\end{table}

\subsection{The Impact of Cultural Adaptation}
We compared how models perform on culturally adapted examples versus those human translated examples that did not undergo such adaptation. During the human translation process, 53 schemas were thoroughly adapted by editing the content of the schema to reflect cultural relevance. 
We observe that, across all models, accuracies are on average 6\% higher on the human translated subset that was culturally adapted (Table \ref{tab:subsets_results}, \textit{Culturally adapted} samples) compared to the human translated subset that did not undergo such adaptation (Table \ref{tab:subsets_results}, \textit{Semantically comparable }samples).
Regardless of the slightly higher accuracies, we do not interpret this as evidence of the models’ stronger ability to reason based on cultural knowledge. This is because the nature of the adaptations varies: some introduce contextually relevant information---such as geographic references---that require cultural or situational understanding, while others involve changes unrelated to reasoning (Table \ref{tab:cultural_adaptation_examples}).

\begin{table}[t]
    \centering
    \small
    \setlength{\tabcolsep}{0.65em}
    \begin{tabular}{lrrrr}
        \toprule
        \multirow{2}{*}{\textbf{Model}} & 
        \multicolumn{2}{c}{\textbf{MT Simple}} & 
        \multicolumn{2}{c}{\textbf{MT Detailed}} \\
        \cmidrule(lr){2-3} \cmidrule(lr){4-5}
        & \textbf{MT} & \textbf{HT} & \textbf{MT} & \textbf{HT} \\
        \midrule
        \multicolumn{1}{l}{\textbf{Altered meaning}} &
        \multicolumn{2}{c}{$n=246$} &
        \multicolumn{2}{c}{$n=191$} \\
        \midrule

        Gemma 3-27B & 63.4 & 77.2 & 62.3 & 74.9 \\
        Qwen 2.5-72B & 67.1 & 72.0 & 65.5 & 69.6 \\
        Llama 3.3-70B & 67.5 & 80.1 & 65.5 & 75.4 \\
        Llama 3.1-405B & 66.3 & 82.1 & 63.9 & 73.8 \\
        \cmidrule(lr){1-5}
        Average & 66.1 & 77.9 & 64.3 & 73.4 \\
        \midrule
        \midrule
        \multicolumn{1}{l}{\textbf{Retained meaning}} &
        \multicolumn{2}{c}{$n=1371$} &
        \multicolumn{2}{c}{$n=1426$} \\
        \midrule
        Gemma 3-27B & 71.2 & 74.1 & 72.1 & 74.5 \\
        Qwen 2.5-72B & 66.5 & 71.1 & 67.5 & 71.5 \\
        Llama 3.3-70B & 70.8 & 73.6 & 71.1 & 74.5 \\
        Llama 3.1-405B & 74.9 & 77.5 & 75.3 & 78.8 \\
        \cmidrule(lr){1-5}
        Average & 70.8 & 74.1 & 71.5 & 74.8 \\
        \bottomrule
    \end{tabular}
    \caption{Comparison of the accuracy scores for schemas where machine translation (MT) either altered or retained the original meaning. 
    }
    \label{tab:meaning-accuracy-compact}
\end{table}

\subsection{The Impact of Incorrect Schemas}
Problematic English schemas were identified manually during the translation process. The results presented in Table~\ref{tab:subsets_results} (\textit {Corrected} samples) confirm the lower quality of these English schemas, as the performance on this English subset is close to the chance level. 

Because these schemas were corrected during the human translation process, performance on the human translated Estonian samples is considerably higher compared to the same subset in English. This improvement in accuracies highlights the importance of quality assurance in schema translation. Importantly, the results indicate that the adjustments preserved the integrity of the tasks and schemas were not rendered trivial by the edits, as the outcomes from the subset of corrected schemas do not differ considerably from those of the subset of 1,617 human translated schemas that did not undergo such corrections (Table~\ref{tab:subsets_results}, \textit{ Semantically comparable} samples).

The performance of the machine translated version of this subset is very low, even below the chance level. This demonstrates that the errors in the original English dataset propagate during the machine translation process in unexpected ways, potentially amplifying the original errors.

\begin{table*}[ht]
\centering
\begin{tabularx}{\textwidth}{lZcc}
\toprule
\textbf{Version} & \textbf{Sentence} & \textbf{Option 1} & \textbf{Option 2} \\
\midrule
EN & So \_ was sorry because Christine's cat was bitten by Jessica's dog when they got into a fight. & Christine & Jessica \\
MT-detailed & Nii et \_ oli kahju, sest Kristiina kass hammustas Jessika koera, kui nad kaklesid. & Kristiina & Jessika \\
Back translation & So \_ was upset, because Kristiina’s cat bit Jessica’s dog when they were fighting. & Kristiina & Jessica \\
\bottomrule
\end{tabularx}
\caption{Example showing original English sentence, its machine-translated version in Estonian, and the back translation with Open AI GPT-5. The schema has become ambiguous in translation 
and the meaning has also shifted (e.g., ``the cat bit the dog'' versus ``the dog bit the cat''). 
}
\label{tab:ambiguous_and_sematic_shift}
\end{table*}

\subsection{The Impact of Machine Translation} \label{subsec:MT_Impact}

To study the effect of machine translation on the results, we had a qualified translator (the same person who did the human translation) manually analyze the subset of 1,617 semantically comparable samples to identify instances where the machine translation had altered the meaning of the original sentence. In this labeling, meaning was the sole evaluation criterion; grammatical issues, unnatural phrasing, or other surface-level errors were not considered problematic unless they affected the sentence’s meaning. The guiding question was: \textit{Does the machine-translated schema convey the same meaning as the original English schema?}.\footnote{These annotations will be released with the dataset.}

The analysis of the machine translations generated with the simple prompt identified 246 (15.2\%) schemas in which the meaning was lost or altered. The translations produced with the detailed prompt resulted in 191 (11.8\%) such schemas. An example of a schema with a shifted meaning is shown in Table~\ref{tab:ambiguous_and_sematic_shift}.

We then compared results on the subset of schemas where machine translation had altered the meaning to those where machine translation had retained the meaning (\textbf{MT} columns in Table \ref{tab:meaning-accuracy-compact}). The shifts in meaning have a clear impact on performance---with one exception (Qwen 2.5-72B), all models perform worse on machine translated schemas with altered meaning. With the simple prompt, the difference in accuracies between the averages of the two subsets is 4.7\% and with the detailed prompt it is 7.2\%. However, it must be noted that, besides human-perceived shift in meaning, machine translated instances contain other artifacts that may affect the accuracy even when the meaning is retained. Thus, we also compared the subset of schemas with retained meaning with the human translated version of the same subset (rows in the lower part of Table \ref{tab:meaning-accuracy-compact}). This comparison shows that the accuracies on the machine translated datasets are lower (3.8\% with the simple prompt and 3.3\% with the detailed prompt) indicating that there are issues beyond altered meaning in the machine translated data that also potentially affect task integrity.

\section{Discussion}

In this paper, we presented the Estonian translation of the WinoGrande test set, which has been translated and culturally adapted by translation specialists and annotated to ensure that translated Estonian schemas satisfy the requirements of the WinoGrande benchmark. We evaluated model performance on the Estonian dataset to assess their reasoning abilities in Estonian. We also explored whether the insights gained from the translation process can be used to engineer a detailed machine translation prompt capable of producing a machine translated dataset of comparable quality. \par 

We observed that the performance for proprietary models was notably good for all models on the English and Estonian dataset, and the results were similar on both datasets. The difference was more pronounced for the open models. The open models performed slightly worse on the Estonian dataset. Smaller range models demonstrated generally poor performance on the task itself, with the exception of EuroLLM-9B, which is specifically trained for European languages. 

During the translation process, we identified a number of flawed schemas in the original dataset. Unlike the original Winograd Schema Challenge dataset, which was developed by experts, WinoGrande was crowd-sourced. This difference in methodology likely accounts for the variability in schema quality and also speaks for the importance of involving language specialists in the development of benchmark tasks requiring advanced linguistic competence. 

We noted a substantial difference in the number of items the annotators labeled as undecidable. Although the discrepancy is numerically large, we do not consider it significant, as the annotators’ accuracy remained close to the reported human performance level (94\%) on the English test set \cite{sakaguchi2019winograndeadversarialwinogradschema}. 

Our experiments with machine translation showed that, despite fine-tuning the translation prompt to account for the dataset’s intricacies and the specific target language, the detailed prompt failed to produce a dataset of human-comparable quality.
 It is also important to highlight that constructing an effective prompt tailored to a specific language and dataset demands detailed knowledge of the dataset's structure and purpose as well as the potential linguistic challenges it presents in a given language. As such, the process still requires significant input from language experts to ensure the prompt is both linguistically and contextually appropriate. 
 Our analysis of the machine-translated dataset focused solely on shifts in meaning during translation. We did not evaluate the translations for grammatical accuracy or the naturalness of the language. Further research could explore how these factors influence model performance.

 \section{Conclusion}

The Estonian translation of the WinoGrande test set provides a valuable linguistic benchmark for Estonian and supports the development and evaluation of multilingual language models. The experimental results demonstrate strong performance from proprietary language models in Estonian, while outcomes from open-source models show greater variability. 
The comparison with machine translation shows that prompt engineering offers limited benefit to translation quality and the results obtained with machine translation may not accurately reflect the models' language comprehension or reasoning. When a model is presented with incoherent or semantically distorted input alongside answer options, it will return a prediction. However, such predictions can be uninterpretable because they do not provide meaningful insight into the model’s linguistic understanding or reasoning capabilities, since the input lacks a coherent structure to support such inference. 

\section*{Acknowledgements}
This work was supported by the National Program for Estonian Language Technology Program (project EKTB104) funded by the Estonian Ministry
of Education and Research, and partially supported
by the Estonian Research Council Grant PSG721.


\section{References}

\bibliographystyle{lrec2026-natbib}
\bibliography{lrec2026-example}

\bibliographystylelanguageresource{lrec2026-natbib}
\bibliographylanguageresource{languageresource}

\appendix

\newmdenv[
  linewidth=1pt,
  roundcorner=5pt,
  backgroundcolor=gray!10,
  linecolor=gray!50,
  innertopmargin=10pt,
  innerbottommargin=10pt,
  innerleftmargin=10pt,
  innerrightmargin=10pt,
]{promptbox}

\twocolumn[

\section{Translation Prompts}

\subsection{Simple Translation}
\label{subsec:simple_p}

\begin{promptbox}
    Please translate the given example in JSON format into Estonian.
    Retain the JSON structure.
    Do not translate the keys.
    Translate the values only.
    Please output only the JSON object, nothing else.
\end{promptbox}

\subsection{Detailed Prompt for Single Items}
\label{subsec:detailed_p}
\begin{promptbox}

Please translate the given example in JSON format into Estonian so that it reads as if originally written in Estonian: natural, fluent, and culturally appropriate. It consists of a sentence and two answer options. \\

Do not translate word for word. Always choose the contextually correct meaning of each word, especially adjectives, idioms, and figurative expressions - never use a literal dictionary equivalent if it does not fit the situation. Ensure that both answer options remain grammatically and semantically valid in the sentence, without revealing or implying which is correct. Pay particular attention to number and case: both answer options must agree in grammatical form so they can fit the same syntactic slot. If the two answer options translations differ in grammatical number (singular vs plural), replace one option with a contextually appropriate alternative so that both options agree in form (e.g. both plural nouns, both singular nouns). The replacement must still preserve the fairness and neutrality of the question. \\

Replace all names with appropriate ones from the provided table, matching gender and applying the correct Estonian case endings. Select appropriate names at random. A name may be reused across different sentences, but not within the same sentence. Maintain neutrality and difficulty: the question must stay fair, challenging, and unbiased. \\

Retain the JSON structure. Do not translate the keys. Translate the values only. Please output only the JSON object, nothing else. Do not escape with backticks or add line breaks.
\end{promptbox}

\vspace{0.5em}

]

\twocolumn[
\subsection{Translation Prompt for Twin Sentences}
\label{subsec:detailed_twin_p}
\begin{promptbox}
The following are sentence pairs with a gap expressed with the underscore and two answer options. Your task is to translate each sentence pair into Estonian so that it
    reads as if originally written in Estonian: natural, fluent, and culturally appropriate. \\
    
    The sentence pairs have a lexical overlap of at least 70\%, and their translations must also maintain at least 70\% overlap. This overlap should be preserved consistently, without introducing unnecessary variation. Do not vary style, synonyms, or sentence structure between them. Answer options must also
    remain exactly the same in these sentence pairs. \\
    
    Do not translate word for word. Always choose the contextually correct meaning of each word, especially adjectives, idioms,
    and figurative expressions — never use a literal dictionary equivalent if it does not fit the situation. Ensure that both answer options remain grammatically and semantically valid in
    the sentences, without revealing or implying which is correct. Pay particular attention to number and case: both answer options must agree in grammatical form so they can fit the same
    syntactic slot. \\
    
    If the two answer options translations differ in grammatical number (singular vs plural), replace one option with a contextually appropriate alternative so that both
    options agree in form (e.g. both plural nouns, both singular nouns). The replacement must still preserve the fairness and neutrality of the question. \\ 
    
    Do not fill the gap.
    Replace all names with randomly chosen ones from the provided table, matching gender and applying correct Estonian case endings. Both names in a sentence must be the same gender,
    and names may be reused across sentence pairs but not within the same sentence. For sentence pairs, use the same names consistently in both sentences. \\
    
    Do not add names where none exist. Maintain neutrality and difficulty: the question must stay fair, challenging, and unbiased.
    You will receive a JSON list containing two examples as input. Your output should also be JSON list containing the translated sentences and answer options. Make sure you translate only
    the values and not the keys. Please output only the JSON list, nothing else. Do not escape with backticks or add addtional line breaks.
\end{promptbox}
]

\end{document}